\documentclass{article}
\usepackage{ijcai25}

\usepackage{times}
\usepackage{soul}
\usepackage{url}
\usepackage[hidelinks]{hyperref}
\usepackage[utf8]{inputenc}
\usepackage[small]{caption}
\usepackage{graphicx}
\usepackage{amsmath}
\usepackage{amsthm}
\usepackage{booktabs}
\usepackage{algorithm}
\usepackage{algorithmic}
\usepackage[switch]{lineno}
\usepackage{tabularx}
\usepackage{natbib}
\usepackage{multirow}
\usepackage{subcaption}
\usepackage{xcolor, colortbl}
\usepackage{microtype} 

\definecolor{lightgray}{gray}{0.9}

\title{DIRIGENt: End-To-End Robotic Imitation of Human Demonstrations based on a Diffusion Model}

\author{
Josua Spisak
\and
Matthias Kerzel\and
Stefan Wermter \\
\affiliations
Knowledge Technology (WTM), University of Hamburg\\
\emails
spisak@uni-hamburg.de
}

\begin{document}

\maketitle

\begin{abstract}
There has been substantial progress in humanoid robots, with new skills continuously being taught, ranging from navigation to manipulation. While these abilities may seem impressive, the teaching methods often remain inefficient. To enhance the process of teaching robots, we propose leveraging a mechanism effectively used by humans: teaching by demonstrating. In this paper, we introduce DIRIGENt (DIrect Robotic Imitation GENeration model), a novel end-to-end diffusion approach that directly generates joint values from observing human demonstrations, enabling a robot to imitate these actions without any existing mapping between it and humans. We create a dataset in which humans imitate a robot and then use this collected data to train a diffusion model that enables a robot to imitate humans. The following three aspects are the core of our contribution. First is our novel dataset with natural pairs between human and robot poses, allowing our approach to imitate humans accurately despite the gap between their anatomies. Second, the diffusion input to our model alleviates the challenge of redundant joint configurations, limiting the search space. And finally, our end-to-end architecture from perception to action leads to an improved learning capability. Through our experimental analysis, we show that combining these three aspects allows DIRIGENt to outperform existing state-of-the-art approaches in the field of generating joint values from RGB images.~\footnote{Code and Data will be made available. The authors gratefully acknowledge support from the DFG (MoReSpace), and the European Commission (TRAIL, TERAIS)}

\end{abstract}

\section{Introduction}
When infants learn how to act, their actions are deeply coupled with their perception, improving their ability to learn by imitation. Despite not being able to see their own faces or bodies, they match gestures or facial expressions shown to them. They can map observed behaviour to themselves, despite anatomical differences~\citep{meltzoff2009foundations}.
\begin{figure}[t]
    \centering
    \includegraphics[width=\columnwidth]{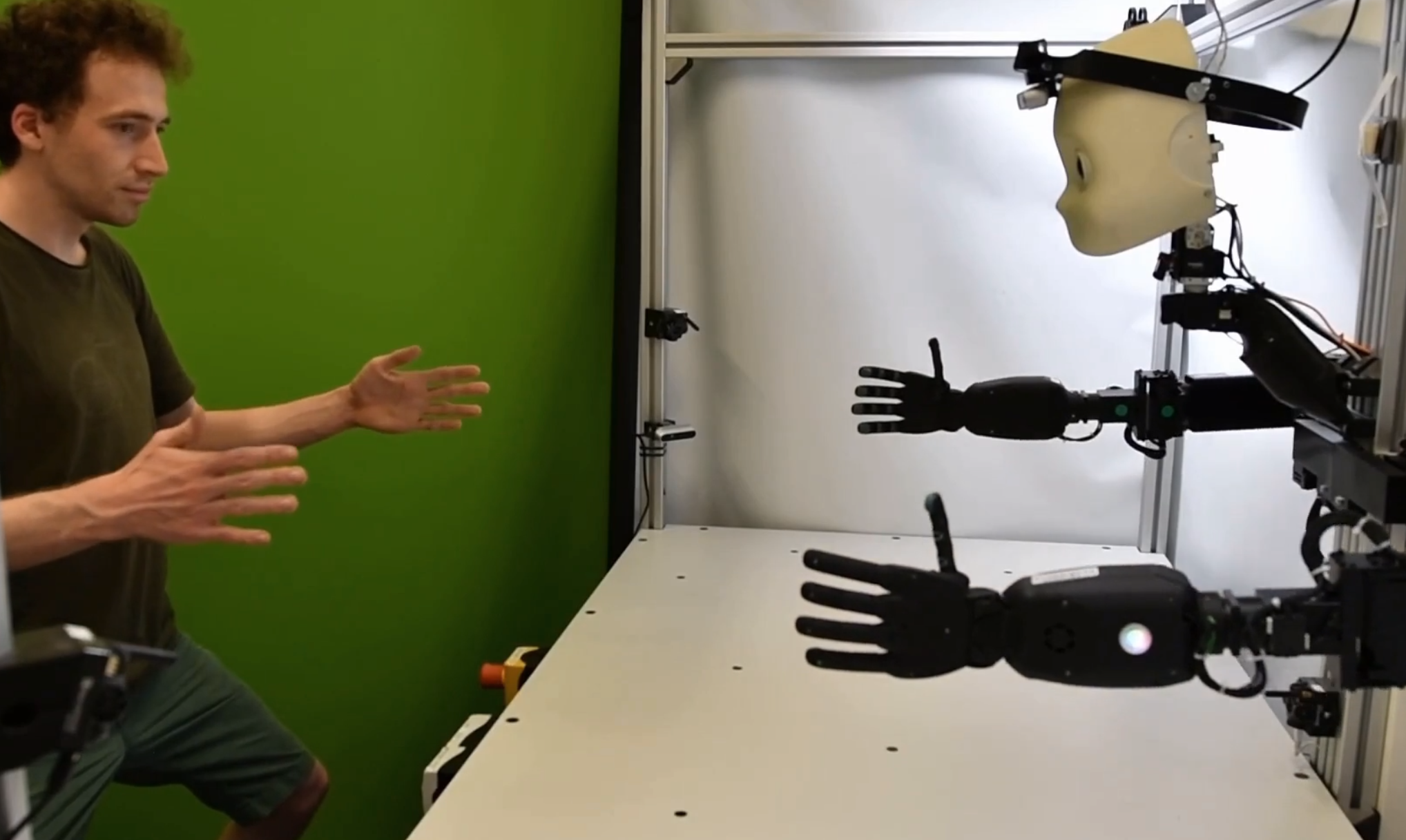}
    \caption{The robot NICOL is on the right side with a human demonstrator on the left side. The robot matches the human's arm poses.}\label{exPicHumToROB}
    \vspace{-10pt}
\end{figure}
\\

Imitating observed behaviour has received significant attention in the field of robotics since it holds the potential for efficient learning of new skills~\citep{finn2017one}. But where infants have a deep coupling between perception and action \citep{meltzoff2009foundations}, many of the robotic approaches learning from human demonstrations have separate modules for perception and action \citep{he2024learning}. With the mapping between the human and the robot often being predefined, which can lead to unnatural behaviour in the robot due to the different anatomies. Learning the actions and perceiving the demonstrated action are processes that we hypothesize to be synergistic and to be improved by learning them jointly.
\\
\\
The individual aspects of both perceiving demonstrations with a robot or doing an imitation are well-researched topics. The advances in processing visual data acquired from optic sensors have led, among other things, to powerful pose detectors that can produce accurate representations of human poses through key points that form a skeleton out of joints. These are typically two-dimensional, corresponding to the two-dimensional images, but approaches exist that allow for three-dimensional pose estimation through lifting from two-dimensional images~\citep{zhu2023motionbert}. Robotic actions are integral to many robotic approaches, and adopting specific joint configurations is a fundamental ability of most robots. The ability to perceive poses and assume joint configurations can be brought together in a mapping between human and robot joints~\citep{zhan2022imitation}.
\\
\\
In this paper, we explore the advantages of developing a deeper coupling between perception and action in learning from observations. To this end, we have built a novel approach that works end-to-end using an RGB image from a human demonstration as input and directly outputting the joint configurations to imitate the demonstration. We collected a dataset consisting of video data from human demonstrations and corresponding robot poses by having humans imitate the robot, as shown in Fig.~\ref{exPicHumToROB} to support the learning of this process since there is no fitting dataset yet. Our model is a diffusion model that takes a single frame from the collected demonstration as its condition. The model generates values for each of the joints of the robot arms. The original poses of the robot that the human imitates are used as the ground truth. Using a diffusion model where the input includes the desired output with added noise during training allows the model to quickly learn the necessary joint values.
\\
Our approach allows the robot to rapidly learn how to imitate a human's arm movements. Benefitting many tasks and applications from teaching gestures to a robot to gaining a deeper understanding of how learning from observation can be explained in robotics.
\\
\\
The main contributions of our paper are:
\begin{itemize}
    \item A novel end-to-end diffusion approach that matches a robot's action to actions demonstrated by a human.
    \item A detailed evaluation of our approach as well as a comparison to a state-of-the-art approach.
    \item A dataset with matching human and robot poses.
\end{itemize}
\section{Related Work}
Imitation learning describes a multitude of related but clearly diverging directions of learning from demonstrations. From the imitation learning observed in nature, different paradigms have been developed in machine learning. Imitation learning can be a subcategory of reinforcement learning (RL) in which demonstrations are used to limit the amount of exploration needed~\citep{hussein2017imitation}. Despite showing some potential of using demonstrations for learning, these approaches are not the focus of our research, as we are interested in the general process of imitation rather than the specific use in RL. Though, our work could be used in RL approaches by providing robotic movements that match human demonstrations, simplifying the use of demonstrations for robotic learning~\citep{taylor2011integrating}. 
\\

Human demonstrations are also used in some approaches outside of RL. In these approaches, the demonstrated actions are commonly used to directly learn something instead of limiting the amount of trial and error required. When imitating, we can either focus on copying the goals of the demonstrator. For example, should the demonstrator fill a cup by pouring water into it from a bottle, whether the demonstrator holds the bottle to the left or right of the cup might not matter~\citep{sermanet2018time, liu2018imitation}. This can be done through a combination of demonstration videos and finding common behaviours, such as the goals of the demonstrations. Alternatively, a single demonstration can be used to deduct the specific task and the high-level actions used to achieve it~\citep{sharma2019third, bahl2022human, spisak2024robotic, yu2018one}.
\\

In contrast, some approaches focus on how the demonstrator moves and how to copy the complete movement. The first step towards this direct matching between the movements of humans and robots is perceiving human motions. This perception is often handled separately through methods such as pose detectors~\citep{zhan2022imitation}, 3D cameras such as the Kinect cameras~\citep{zhang2019real, ou2015real}, motion capture systems~\citep{stanton2012teleoperation, koenemann2014real} and combinations of  methods~\citep{do2008imitation}. These human pose representations are used in further processing steps. The representation is often an abstraction of a human, such as a stick figure or even just the position of the end effectors, keeping all relevant information needed for the imitation while removing possible distractions from high-dimensional source data.
The correspondence problem describes the challenge of creating a mapping between a human and a robotic body~\citep{nehaniv2002correspondence}. Many robots diverge from the human kinematic chain, making a direct one-to-one mapping impossible. Moreover, the speed of movement also has to be adjusted, with robots often moving slower than humans. While partial mappings can be learned from a group of joints to a single actuator, a direct one-to-one mapping is problematic.~\citep{stanton2012teleoperation}.
\\
The correspondence problem can also be considered an optimisation problem, where the difference between the demonstrator's and the robot's joints have to be minimised~\citep{suleiman2008human}. It can be simplified by focusing on specific parts of the body instead of the complete kinematic chain~\citep{koenemann2014real}. To completely avoid the correspondence problem of mapping from human to robot, a human demonstration can be translated to look like robot demonstrations~\citep{smith2019avid}, changing it from a mapping between joints to a visual mapping. Once the joint configuration of the demonstrator has been mapped onto the robot, the imitation needs to be executed. During this, further limitations to the movements come into play, such as self-collision~\citep{arduengo2021human}.
\\
\\
In summary, existing approaches tend to decouple perception from the action, whether this means first analysing the goals and using them to define the task or first analysing the movement before finding a mapping. We hypothesize that the right integration of perception and action will improve the imitation by allowing for a more natural mapping between the perceived demonstrator and the acting imitator.

\section{Approach}\label{Approach}
Rare existing datasets such as the MIME~\citep{sharma2018multiple} or  RH20T~\citep{fang2023rh20t}  provide matching pairs of human and robot demonstrations. However, while the demonstrations match in their effect, for instance plugging in a cable or sliding an object to the side, the actions do not match. Since robots often move at different speeds and have different morphologies, recording one-to-one matches between human and robot demonstrations can be challenging. This exact match between the recorded joint values and the third-person recording of the demonstrations is, however, the data needed for imitation learning. For this purpose, we recorded a new dataset to address this lack of existing datasets.
\\

\begin{figure}[t]
    \centering
    \includegraphics[width=0.49\columnwidth]{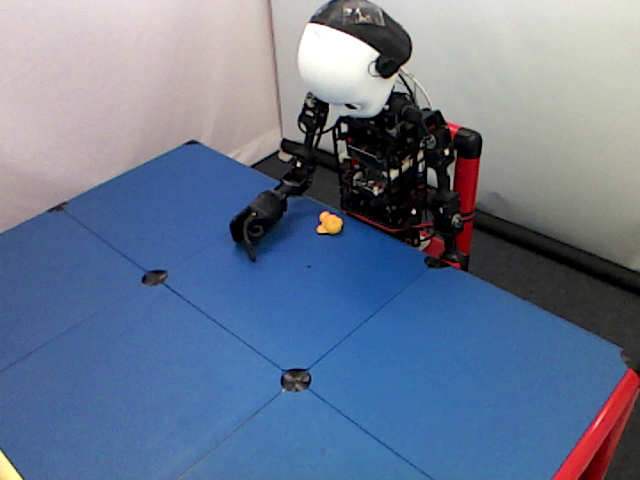}
        \label{EMILex}
    \includegraphics[width=0.49\columnwidth]{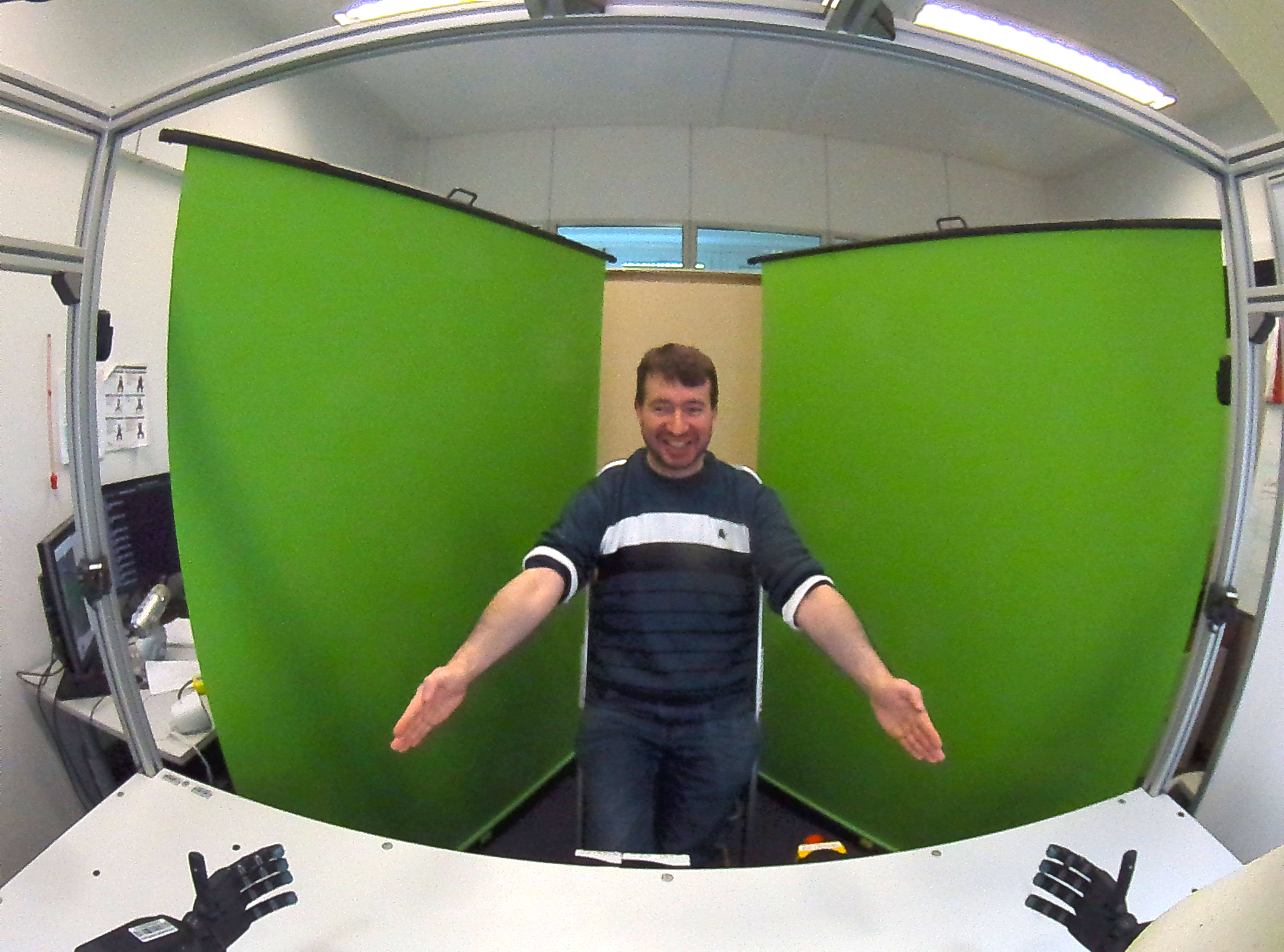}
        \label{DirEx}
    \caption{Comparison of datasets: on the left is the EMIL dataset and on the right is the Direct Imitation Dataset.}
    \label{fig:combined}
\end{figure}
\textbf{DIRI (DIrect Robotic Imitation) dataset.} Since humans are capable of imitation, we reverse the task during the collection of the dataset, having the human participants imitate the robot. 
Through this, we can record an intuitive match between human and robot morphology. We make the imitation as easy as possible to improve the quality of that matching process. The robot moves very slowly and only changes directions whenever it reaches the end of its movement constraints. This makes its movement as predictable as possible for the human participant. To diversify the dataset without making it harder to imitate, we record three separate imitations. In each, the robot starts at a randomly chosen position inside its movement constraints and, from there, starts to move along the three axes (left or right, up or downwards, forwards or backwards) at randomly chosen speeds. This leads to each of the runs containing many previously unknown joint configurations as well as unseen human imitations. Further diversity in the data is gained by recording ten participants. The dataset contains the RGB images recorded through the camera inside the left eye of our robot NICOL~\citep{kerzel2023nicol} as shown in Fig.~\ref{fig:combined}, as well as the joint values of both of its arms. For each arm, there are 13 joint values, of which the last five relate to its hands. The images were saved with a resolution of 256 x 256 pixels. Since the recording rate of the RGB images differed from the recording rate of the joint values, both information paths were synchronised during post-processing. For each of the recordings, the robot was steered towards 600 separate positions. Its movement was controlled through CycleIK~\citep{habekost2023cycleik} and directed by three-dimensional positions chosen for its left and right end effectors. Since the CycleIK is not deterministic in finding joint configurations for a given three-dimensional end effector position, a variety of joint configurations exists in the recorded dataset. The recording generally lasted between three and five minutes, as the distance between the positions could vary between runs. While the robot was only steered towards 600 positions in each recording, the sampling rate was higher leading to approximately 110000 recorded RGB images of human demonstrations and corresponding joint values for the robot demonstrations. \\

\textbf{EMIL dataset.} The EMIL dataset~\citep{heinrich2018embodied} contains continuous, multi-modal and body-rational data. This includes image recordings of the NICO robot~\citep{kerzel2017nico} from an external camera as shown in Fig.~\ref{fig:combined}, and the recorded joint values for its right arm. The dataset also provides information on which joint values belong to which of the images. These characteristics allow us to use it for our approach of generating joint values depending on third-person recordings. The data was recorded in a setup, where NICO is seated in a chair next to a table and performs four different object manipulation tasks. In total, there were 30 recordings for each of the tasks resulting in 59500 samples. \\

\textbf{Benefits of two datasets.} The two datasets we use differ in some key aspects, allowing us to analyse the strengths and limitations of our approach. In the EMIL dataset, our approach does not need to overcome the correspondence problem, human errors or different demonstrators in the EMIL dataset. For the direct imitation dataset, the NICOL robot has more degrees of freedom (DoF) than the NICO robot, making the generation of joint values more complex for the direct imitation dataset. In contrast, the EMIL dataset allows us to test how well the approach generalises over new tasks and how independent it is from a given dataset.
\subsection{DIRIGENt}
\begin{figure*}[t]
    \centering
    \includegraphics[width=\textwidth]{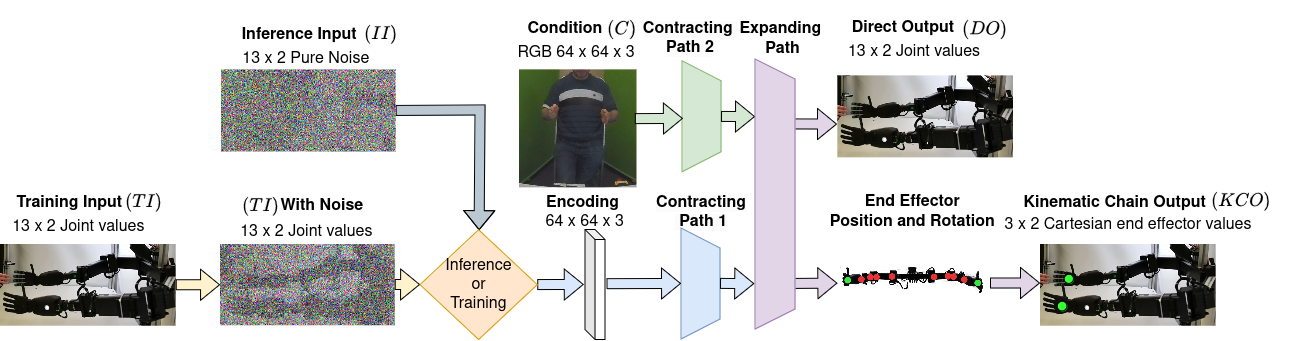}
    \caption{Our architecture. On the left side the differences between training and inference are shown as well as all inputs to the model, on the right side the neural architecture and the outputs are shown.}\label{arch}
\end{figure*}
To accomplish this task of imitating human behaviour our approach has to reverse the imitation of the humans. To do this successfully, our approach needs to approximate the following function: 
\begin{equation*}\label{eq:1}
    \underbrace{D_M(D_C(}_{\text{robotic imitation}}\underbrace{P_A(P_M(P_P(}_{\text{human imitation}}
    \underbrace{D_{TI})))))}_{\text{robotic demonstration}}
\end{equation*}
\(D_{TI}\) refers to the robotic action that participants imitated during dataset collection, and we assume it is accurate for our training input. Next, \(P_P\) represents the participant's perception of \(D_{TI}\), which may introduce some noise. Followed by \(P_M\), the mapping from the robot to the participant's body, and \(P_A\), the participant's action, which can add further noise. After the participant's action, the model perceives the participant, denoted as \(D_C\), which serves as the condition in our approach. Finally, the model must map \(D_M\) from the participant's body back to the robot to generate the joint configuration \(D_{DO}\) that imitates the human. 
To direct the learning in our approach we devise the loss term:
\begin{equation*} 
    \mathbf{L} = \omega_1 \cdot \text{MSE}(D_{DO},D_{TI})+\omega_2 \cdot \text{MSE}(D_{DO_{(eef)}},D_{TI_{(eef)}})
\end{equation*}
Where the shown function needs to be modelled correctly to optimize $\mathbf{L}$ which reduces the distance between the demonstrated and imitated joint configurations while emphasizing the end effector positions ($eef$). The two mean square errors (MSE) are weighted by $\omega_1$ and $\omega_2$. With this in mind, we propose our DIRIGENt approach that does not merely focus on the $D_P$ and $D_M$ to imitate humans through robots, but rather uses both $D_{TI}$ and $P_A$ during training to create an accurate model for the proposed problem.
\\

\textbf{Basic structure.} Our model is composed of two contracting paths, one expanding path and one kinematic chain for the forward kinematics shown in Fig.~\ref{arch}. The first of the contracting paths starts by encoding the 26 (2 x 13) joint values (TI or II) and reordering them to create an array with the dimensions of 3 x 64 x  64 to match the dimensions of the RGB image from the condition (C). This allows us to concatenate both information streams later during the expanding path. The RGB image for the condition (C) goes through the second contracting path. Both paths proceed with a series of encoding blocks that consist of convolutional layers enhanced through multi-head attention. In these blocks, the dimensions of the arrays are down-sampled from 64 x 64 x 3 to 16 x 16 x 128 through max-pooling layers. The reverse happens in the expanding path, where up-sampling layers are used to increase the dimensionality of the sample to 64 x 64 x 64 before ending in a linear layer that decodes the data to end with one array for the 26 joint values (DO). Both contracting paths are connected to the expanding path, the first contracting path leads into two double convolutional blocks and then into the first up-sampling block of the expanding path. The second contracting path is connected between the self-attention blocks of the contracting path and the up-sampling blocks of the expanding path. By connecting the second contracting path in multiple resolutions, more spatial information is kept. The contracting and expanding paths are based on the U-net~\citep{ronneberger2015u} which was originally used for image segmentation, and an approach on perspective transfer~\citep{spisak2024diffusing}.

\textbf{Forward kinematics.} The final part of our model starts with a linear layer which reforms the sample to fit the amount of joint values needed, in the case of our direct imitation dataset and the NICOL robot that is 13 joint values for each of the arms. Then, the joint values go through a kinematic chain that produces the three-dimensional coordinates of the end effector (KCO)~\citep{Zhong_PyTorch_Kinematics_2024}. This allows us to optimise the joint error and the Cartesian distance between the produced and the desired position of the end effector. A desired position can be reached through multiple joint configurations because of the high number of DoF that NICOL's arms possess. We use the kinematic chain to have the benefits of both representations of the robot's arm configuration.

\textbf{Diffusion.} Apart from the kinematic chain, we also make use of the benefits of diffusion models to enhance how our model learns to generate joint values. In diffusion models, the input has the same shape as the output. This input differs in the training and inference phases: During training, the input is copied from the desired output, but some amount of noise is added to the input (TI). The network is then trained to either predict the noise that was added, so it can be removed again, or to predict the original desired output. In our model, there are 1000 levels of noise. The level of noise applied at any given training step is decided by a cosine noise schedule, which allows for a smoother transition from low amounts of noise to high amounts of noise~\citep{nichol2021improved}. During inference, the input is always noise (II). If the model has been trained to predict the noise, a number of denoising steps equal to the amount of noise levels are needed in which, step by step, a level of noise is predicted and removed from the input until an image is generated. If the model was trained to predict the desired output, as is the case in our approach, the inference is significantly sped up by reducing the number of denoising steps. Providing the model with a noisy version of the desired output during training can be especially helpful when learning joint values. The challenge of deciding between solutions that are significantly different in the joint space while very similar in the Cartesian space is removed. The second mechanism of the diffusion model is the iterative denoising during inference, which has the potential to improve the results as outputs are refined over multiple steps. To facilitate these two mechanisms, two contracting paths exist: one encodes the noisy input and transfers information about the noise level of any given sample, and the other encodes information about the recorded demonstration. 

\textbf{Condition.} To direct the results of diffusion models, a condition can be used~\citep{choi2021ilvr}. The condition works like a label given to the network and allows it to associate specific outcomes with the label. In our approach, the condition is the recorded frame of a demonstration for which we want to generate fitting joint values to imitate the demonstration.
The condition (C), an RGB image, is encoded through the second contracting path of our model in the same way during the training and inference phase. During inference, the condition is the only information the model can use to generate joint values, making it important that as much information from the condition is kept as possible. 

\textbf{Hyperparameters.} Our model has close to eight million parameters and was trained on commercial GPUs. We used the Adam optimiser, a batch size of six, and trained the model for twenty epochs. The inference takes 10 ms on GeForce RTX 3060, allowing for real time processing.
\section{Results}
\begin{figure}[h]
    \centering
    \includegraphics[width=0.49\columnwidth]{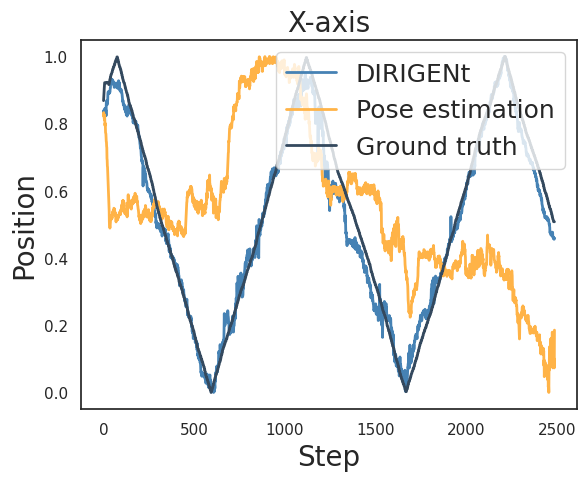}
    \includegraphics[width=0.49\columnwidth]{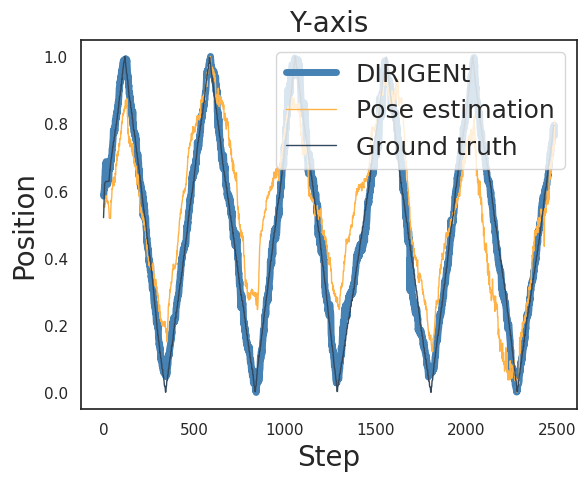}
    \caption{The end effector position in the label, the model's prediction and pose estimation over 2400 frames normalized to be between 0 and 1 along the x-axis and the y-axis.}\label{YaxisEx}
    \vspace{-10pt}
\end{figure}

\begin{table*}[t]
    \centering
    \begin{tabularx}{\textwidth}{>{\bfseries}c  c X X >{\centering\arraybackslash}X >{\centering\arraybackslash}X >{\centering\arraybackslash}X}
        \toprule
        \multirow{2}{*}{Experiment} & \multirow{2}{*}{Configuration} & Joint Loss & Cartesian Loss & \multirow{2}{*}{\shortstack{X-axis \\ (m)}} & \multirow{2}{*}{\shortstack{Y-axis \\ (m)}} & \multirow{2}{*}{\shortstack{Z-axis \\ (m)}} \\
        \midrule
        \textbf{HoRoPose} & \cellcolor{lightgray}DIRI fold 10 &\cellcolor{lightgray}  0.023 &\cellcolor{lightgray}  0.125 &\cellcolor{lightgray}  0.226 &\cellcolor{lightgray}  0.136 &\cellcolor{lightgray} 0.100 \\ 
        \midrule
        \multirow{2}{*}{\shortstack{DIRIGENt (ours) \\  for EMIL} }
        & EMIL Random 90:10 & 0.0007 & 0.005 & 0.003 & 0.004 & 0.004 \\
        & EMIL Average from all tasks & 0.047 & 0.149 & 0.024 & 0.046 & 0.028 \\
        \midrule
        \multirow{2}{*}{\shortstack{Ablation:\\ Pose Estimation} }
        & DIRI fold 10 Pose 2D & 0.014 & 0.113 & 0.068 & 0.064 & 0.057 \\ 
        & DIRI fold 10 Pose 3D & 0.021 & 0.091 & 0.088 & 0.075 & 0.068 \\ 
        \midrule
        \multirow{4}{*}{\shortstack{Ablation:\\ Cartesian vs Joint}} 
        & DIRI fold 10 - Cartesian loss & 0.010 & 0.160 & 0.048 & 0.071 & 0.043 \\ 
        & DIRI fold 10 - Joint loss  & 0.134 & 0.130 & 0.052 & 0.053 & 0.038 \\ 
        & DIRI fold 10 + Cartesian consistency & 0.022 & 0.255 & 0.120 & 0.220 & 0.066 \\ 
        & DIRI fold 10 + Direct Cartesian generation & 0.017 & 0.146 & 0.064 & 0.078 & 0.047 \\ 
        \midrule
        \multirow{2}{*}{Ablation: Diffusion} 
        & DIRI fold 10 - noisy input for training & 0.033 & 0.250 & 0.079 & 0.137 & 0.089 \\ 
        & DIRI fold 10 + 50 iterative denoising steps & \textbf{0.009} & 0.103 & 0.040 & 0.056 & 0.039 \\ 
        \midrule
        \textbf{Ablation: Time} & DIRI fold 10 + overlayed past frame & \textbf{0.009} & 0.100 & 0.041 & \textbf{0.052} & 0.044 \\ 
        \midrule
    
        \multirow{4}{*}{DIRIGENt (ours)} 
        & DIRI fold 10 + 80 epochs & \textbf{0.009} & 0.104 & \textbf{0.039} & 0.054 & \textbf{0.038} \\ 
        & DIRI random 90:10 & \textbf{0.0001} & \textbf{0.0232} & \textbf{0.002} & \textbf{0.003} & \textbf{0.002} \\ 
         &
        \cellcolor{lightgray}DIRI fold 10 &
        \cellcolor{lightgray} 0.010 &
        \cellcolor{lightgray} 0.111 &
        \cellcolor{lightgray} 0.043 & 
        \cellcolor{lightgray}0.060 & 
        \cellcolor{lightgray}0.041 \\ 
        & DIRI average from all folds & 0.010 & 0.106 & 0.048 & 0.061 & 0.043 \\ 
        \bottomrule
    \end{tabularx}
    \caption{The results for our experiments. The reported metrics are the joint loss (MSE), the Cartesian loss (MSE), and the distances along each axis for the end effector position.}
    \label{tab:results}
\end{table*}
\textbf{Evaluating the imitation capability.} To evaluate the performance of our model, we calculate the distance along the three axes x, y, and z between (1) the ground truth which is the position of the end effector that the robot originally showed to the human participant and (2) the position of the end effector generated by the model. We further include the joint loss to compare the different arm poses. The error of the human in imitating the robot is included in this measure. Since the NICOL robot is not only supposed to follow the human movements but also adapt them to its own body, there is no way to avoid the human error when trying to learn human-like imitation, the model manages to adapt to this error as can be seen in Fig. \ref{YaxisEx}. 
We randomly split our dataset into a training set and a test set, with a split of 90:10. The recording setup, where the robot starts in a random position and moves at random intervals along the axes, makes redundant samples in the dataset very unlikely. Therefore, the test set consists of samples that are not in the training set. The total range of movement of the robot is 40 cm on the x-axis, 50 cm on the y-axis and 30 cm along the z-axis. For the 10 \% of the data used as the test set, our approach is, on average, 2 mm away from the desired x-position, 3 mm away from the desired y-position and 2 mm away from the desired z-position, as detailed in Tab. \ref{tab:results}. The relative errors are 0.004, 0.006 and 0.007 respectively. As expected, the x-axis is the easiest to imitate, since the up-down movement range of the robot is completely in the range of every human participant. The y-axis is harder because the robot can reach further due to its long arms, meaning the human imitator will have to adjust and cannot simply mirror the movement. Finally, the z-axis is the hardest, presumably because much of the depth dimension is lost in the 2D recordings.\\

In addition to our internal evaluation, we compare our results with the state-of-the-art approach HoRoPose \citep{ban2025real}. Since their approach was designed to estimate joint values for a robot using an RGB image, it fits our dataset well and shares the purpose of estimating joint configurations of robots. Rather than using any form of diffusion, they utilise a CNN-based network. We trained their approach from scratch using their default hyperparameters for the same 20 epochs as our model. The results of using HoRoPose on our dataset are shown in Tab.~\ref{tab:results}. While the external approach yields better results than our model without the diffusion input, our standard model outperforms HoRoPose.
\\

\begin{figure}[t]
    \centering
    \includegraphics[width=\columnwidth]{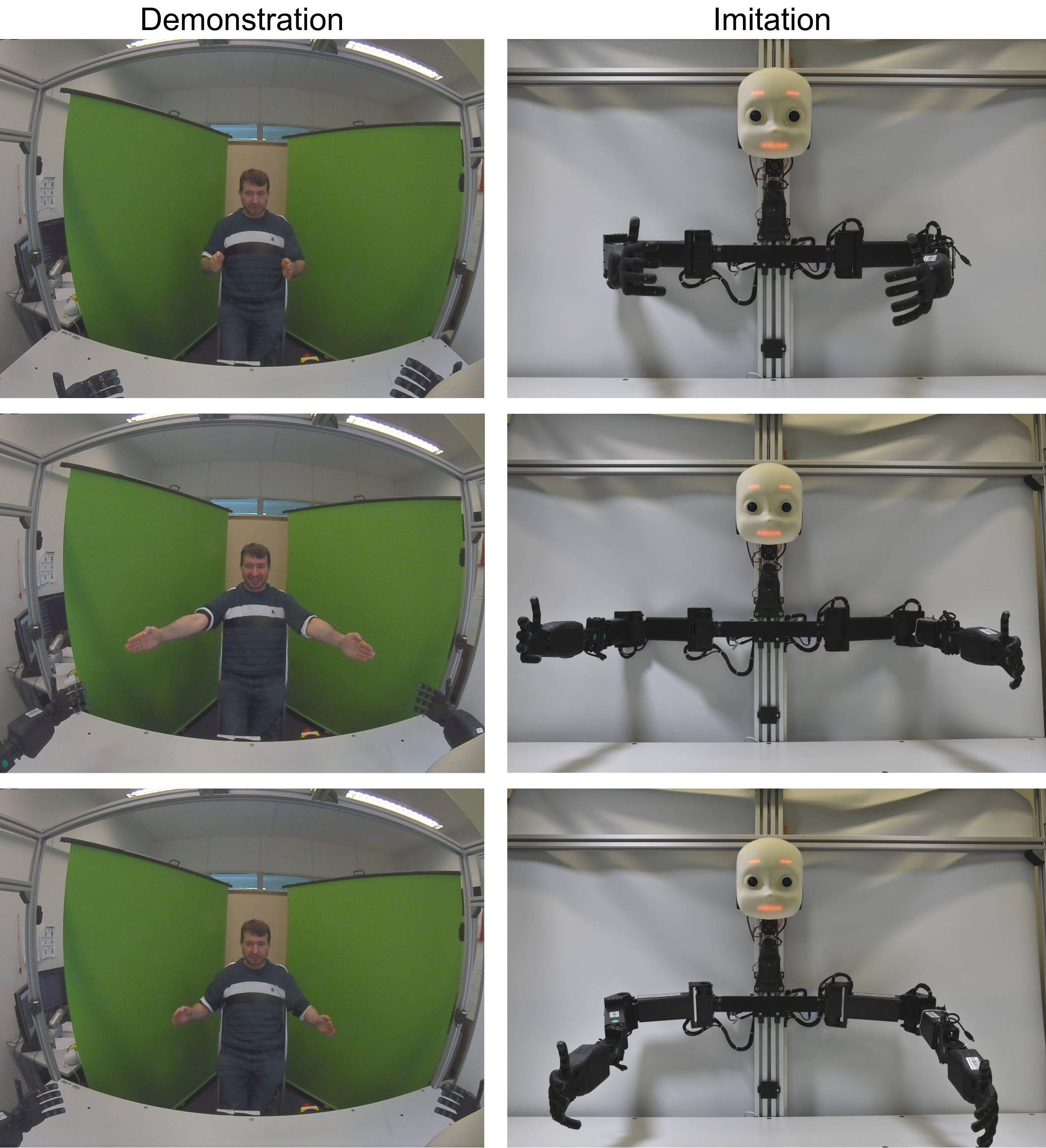}
    \caption{On the left is the human demonstration, and on the right is the robotic imitation based on the generated joint configuration.}\label{fig:qualEx}
    \vspace{-10pt}
\end{figure}

\textbf{Generalisation over humans.} The capability of our approach to generalise over different humans was assessed through a 10-fold cross-validation. For each fold, one of the demonstrators is kept for validation while the model is trained on the other nine. The average distance along the axes increases to 4.8 cm for the x-axis with a standard deviation of 0.37, 6.1 cm for the y-axis with a standard deviation of 0.42 and 4.3 cm with a standard deviation of 0.17 for the z-axis. The increase in error is expected due to the different morphologies among humans. Since the human imitators have to find a way to adjust their body movements to imitate the robot movements, there is a unique mapping between their body and the robot's body, while the approach learns a general mapping from any human to the robot. However, the approach remains capable of imitating the movement patterns of humans unseen during training, as shown in Fig.~\ref{fig:qualEx}.\\

\textbf{Generalisation over tasks (EMIL).} NICO's motion range in the EMIL dataset is 21.89 cm for the x-axis, 31.29 cm for the y-axis and 25.64 cm for the z-axis. When training on a random split of 90:10 of the data, our approach had an average distance of 3 mm for the x-axis, 4 mm for the y-axis and 4 mm for the z-axis on the validation set which is in line with Inverse Kinematic (IK) approaches for the NICO robot \citep{GHW24, habekost2024inverse}. We further tested how well the approach generalises from one of the four tasks (lift, push, pull, scoot) to another. For this, we trained our model on one task and evaluated it with each of the other three tasks. The best generalisation happened between the lift and scoot tasks, where distances close to 1 cm were reached. On average, the generalisation to other tasks had a distance of 2.4 cm, 4.6 cm and 2.8 cm for the x-,y- and z-axis. Despite the tasks containing completely new movements and joint configurations, such as the grasping motions for the lifting, our approach imitates the demonstrated movements accurately. 
\subsection{Ablations}
\textbf{Pose estimation.} To evaluate the advantages of having an end-to-end approach instead of using pose estimation, we change the condition of our model from RGB images to detected poses. If the pose estimations hold all relevant information for the imitation, we would expect there to be no difference in the results. We used the state-of-the-art 2D pose estimator RTMO~\citep{lu2024rtmo} with mmpose~\citep{mmpose2020} to create 2D poses for the DIRI dataset. From these, we created images showing the 2D keypoints with the same dimensions as the original recordings, so that our model architecture can stay the same, allowing for a direct comparison of the different conditions. We split our dataset according to the participants since one advantage of poses could be that it is easier to generalise since much of the differences between humans are abstracted in 2D Skeletons. The data from the last participant was used for testing, while the data recorded from the other participants was used for training. The results are worse across every measurement compared to our base approach, supporting our assumption that an end-to-end approach might learn additional details needed for an accurate imitation. The distances along the x-axis, y-axis and z-axis increased by around 25 mm, 4 mm, and 16 mm respectively. The added error alongside the z-axis is expected since some details in the RGB data, such as distance-dependent hand sizes, are lost. However, while training converged faster with pose estimations, our approach performed better after completing the training.

3D pose estimators are not as common as 2D pose estimators due to the increased difficulty in estimating the third dimension given only a 2D image. However, since 3D information is very much needed to imitate 3D poses, we also evaluate our approach using 3D pose estimations. We use a state-of-the-art 3D pose estimator, motionbert~\citep{zhu2023motionbert}. To generate 3D poses, the motionbert approach uses lifting, which takes the 2D poses and estimates the 3D poses from them. Since the condition of our approach is an array representing a 2D image, we changed the colouration of the skeleton key points depending on their z-axis to preserve the information from the 3D pose estimations. Despite this additional information, our base approach performs better. The distances along the x-, y- and z-axes increased by around 40 mm, 15 mm and 25 mm respectively. We assume that the discrepancies between the accuracy of the 2D and 3D pose estimators lead to better results from the 2D poses.
\\

\textbf{Cartesian and joint loss.} Our approach directly produces joint values from the neural network and uses a kinematic chain to additionally procure the Cartesian position of the end effector, both of these outputs are used in the loss calculation. The purpose of the Cartesian loss is to alleviate the challenge of imitating the complete joint configurations that might have redundancies due to the high DoF of the robot, while still imitating the pose of the arm through the joint loss. To evaluate the use of each of the losses, we run our approach with either only the joint loss or only the Cartesian loss. When only using the Cartesian loss, the distance along the three axes only changes slightly, decreasing for the y- and z-axes while increasing along the x-axis. As expected, the joint loss increases significantly to more than ten times the loss when using both losses, as the approach finds other redundant inverse kinematic solutions to reach the same end effector pose. As the goal of our approach is to imitate the demonstration fully rather than solely the end effector, it becomes clear that the joint loss is essential. On the other hand, solely utilising the joint loss without the Cartesian loss, increases the distance from the desired position along each axis. The combination of both losses outperforms either loss on their own, although the joint loss appears to be more important than the Cartesian loss for our purpose. This is surprising as the Cartesian loss significantly improved the joint generation in similar tasks \citep{GHW24, pavllo2018quaternet, habekost2023cycleik}, we assume that this is a consequence of having the noisy input during training which narrows down the possible choices for the generated joint configuration eliminating some of the challenges stemming from redundant joint configurations.
\\

The importance of the joint loss can be explained by the architecture of our model, in which only the joint values are given to the model during training. Therefore, we evaluate if extending the model to include the Cartesian information from the kinematic chain in the diffusion process is beneficial. Towards this purpose, the seven values describing the position and rotation of the end effector gained from the forward kinematics, are added to the in- and output of our neural network. The model now produces two Cartesian results: the direct output of the model, and the output from the kinematic chain that uses the joint values from the model. The directly generated Cartesian result can be compared to the goal Cartesian value as a consistency measure to the Cartesian output from the model's kinematic chain. However, the added complexity cannot improve our approach, giving weight to the hypothesis that the Cartesian loss is less important to our model because the redundancy for joint configurations is reduced through the diffusion training mechanisms.
\\

\textbf{Diffusion.} To evaluate the effects of using a diffusion model, we change our approach so that during training, the chosen noise level on the input is always the highest possible amount of noise. The condition remains unchanged, essentially making the training process the same as the inference process. However, the model has no information about the desired output during the training, making the first contracting path essentially superfluous. Removing this from our model worsens the results, increasing the distances along the x, y and z-axis respectively by 31 mm, 77 mm and 46 mm, showing just how important the diffusion is to our approach. 
Another aspect of the diffusion models is the iterative denoising, which can be used during inference to go over samples multiple times with different levels of positional encoding for the noise level, potentially allowing the network to adjust to the given noise level. In the reported results so far no iterative denoising has been used, instead, the direct output of the neural network has been evaluated, to evaluate the benefits of iterative denoising to our approach we add 50 denoising steps here. In other diffusion approaches, the iterative denoising can be observed to improve details \citep{liu2023diffusion}, which are refined over the denoising steps. In our approach, the difference between high-level and low-level aspects of the output is less clear, which is one reason why even a single step is enough for a good performance in our model. While in other approaches the model could create the high-level features at higher noise steps and focus on low-level parts towards the end of the denoising \citep{ho2020denoising}, even small differences in our output could result in a much higher loss since the resulting end effector position might be changed significantly. \\

\textbf{Time.} The last aspect of our approach we investigate is the continuity of the demonstrations and, consequently, the imitations. Since the recordings stem from continuous demonstrations, the information about the movements needed to take on a given pose might be beneficial when imitating said pose. Therefore, we evaluate how adding information about these movements can affect our approach. Towards this purpose, we overlayed ten samples before the current frame onto the current frame with an opacity of 0.5 which results in an image that holds information about motion. However, 
the results show that this additional information is not required by our model. It appears that all relevant information needed to imitate a pose can be gained directly from an image of that pose.
\section{Conclusions}
In this paper, we have introduced a novel diffusion-based approach that allows robots to imitate human poses. Additionally, we recorded a dataset uniquely fit for this task of robot learning to imitate humans. We evaluated many conditions of our approach, including how dependent it is on a specific dataset, how it generalises over different humans, and how it generalises over different tasks. We further tested which parts of the approach are essential and which parts could potentially be modified.  The experimental findings demonstrated the approach's efficacy, particularly for known humans, while also exhibiting generalisation capabilities for unknown individuals. The generalisation over tasks was found to be contingent upon the similarity between the tasks. Additionally, the experimental results indicated that RGB images contain more information than two- or three-dimensional pose estimations. Furthermore, the incorporation of label noise as input during training was found to enhance the generation of joint values.
Further improvements to the approach are possible, such as improved generalisation or inclusion of acceleration values. However, we have shown that an end-to-end diffusion model is capable of accurately generating joint values that allow a humanoid robot to imitate humans as it sees them, demonstrating capabilities in understanding human poses and translating them to a humanoid robotic body.
\bibliographystyle{named}
\bibliography{ijcai25}

\end{document}